\documentclass[10pt]{article}

\usepackage{preprint}

\usepackage{inconsolata}
\usepackage{fvextra}
\usepackage[dvipsnames]{xcolor}
\usepackage{graphicx}
\usepackage[export]{adjustbox}
\usepackage{amsmath}
\usepackage{amssymb}
\usepackage{booktabs}
\usepackage[shortlabels,inline]{enumitem}
\usepackage{etoolbox}

\usepackage{iftex}
\ifPDFTeX
  \usepackage[utf8]{inputenc}
  \usepackage[T5,T1]{fontenc}
\else
  \errmessage{Compile with pdfLaTeX only}
\fi

\newcommand{\vn}[1]{{\fontencoding{T5}\selectfont #1}}

\usepackage{multirow}
\usepackage{array}
\usepackage{microtype}
\usepackage{capt-of}

\makeatletter
\let\ax@footnotesize\footnotesize
\newif\ifax@pair
\let\ax@figure\figure \let\ax@endfigure\endfigure
\let\ax@table\table \let\ax@endtable\endtable
\renewenvironment{figure}[1][tbp]{\ifax@pair\captionsetup{type=figure}\else\ax@figure[#1]\fi}{\ifax@pair\else\ax@endfigure\fi}
\newsavebox{\ax@tabbox}
\renewenvironment{table}[1][tbp]{%
  \ifax@pair\captionsetup{type=table}\let\footnotesize\ax@footnotesize\else\ax@table[#1]\fi
  \gdef\ax@tablabels{}\let\ax@label\label
  \setbox\ax@tabbox\vbox\bgroup\hsize=\linewidth
    \renewcommand{\caption}[1]{\gdef\ax@tabcap{##1}}%
    \renewcommand{\label}[1]{\g@addto@macro\ax@tablabels{\ax@label{##1}}}}{%
  \egroup
  \caption{\ax@tabcap}\ax@tablabels
  \unvbox\ax@tabbox
  \ifax@pair\else\ax@endtable\fi}
\renewenvironment{table*}[1][tbp]{\begin{table}[#1]}{\end{table}}
\captionsetup[table]{position=above,skip=5pt}
\AtBeginEnvironment{tabular}{\ifax@pair\setlength{\tabcolsep}{3.5pt}\fi}
\newcommand{\float}[1]{\@ifundefined{ax@float@#1}{\@latex@error{No placement defined for float #1}\@ehc}{\csname ax@float@#1\endcsname}}
\newcommand{\ax@pair}[5][\vspace{0pt}]{%
  \ax@figure[t]\ax@pairtrue
  \begin{minipage}[t]{#2\linewidth}#1#3\end{minipage}\hfill
  \begin{minipage}[t]{#4\linewidth}#1#5\end{minipage}%
  \ax@pairfalse\ax@endfigure}
\@namedef{ax@float@fig_grade}{\ax@pair{0.49}{\begin{figure}[t]
\centering
\includegraphics[max width=\linewidth]{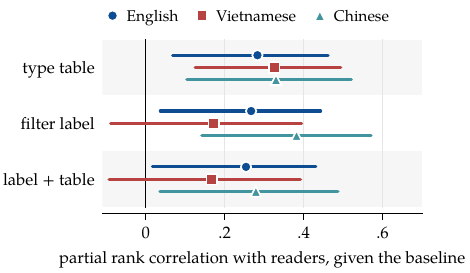}
\caption{\textbf{Severity carries information that simpler charges miss.} Each row is the severity's partial correlation with readers' ratings within severity band, given the baseline at left. \emph{Label + table} is the filter label and the type table together.}
\label{fig:grade}
\end{figure}
}{0.48}{\begin{table}[t]
\centering\footnotesize
\setlength{\tabcolsep}{8pt}
\begin{tabular}{@{}lccc@{}}
\toprule
 & English & Vietnamese & Chinese \\
\midrule
\multicolumn{4}{@{}l}{\emph{Preference pool}} \\
\quad \semcer{} & 0.71\rlap{$^{*}$} & 0.79 & 0.78\rlap{$^{*}$} \\
\quad \cfcer{}  & 0.58 & 0.80 & 0.60 \\
\midrule
\multicolumn{4}{@{}l}{\emph{Normalization-proof pairs}} \\
\quad \semcer{} & 0.77\rlap{$^{*}$} & 0.53 & 0.83\rlap{$^{*}$} \\
\quad \cfcer{}  & 0.09 & 0.50 & 0.47 \\
\bottomrule
\end{tabular}
\caption{\textbf{On pairs drawn where \cer{} and \semcer{} disagreed, readers mostly side with \semcer{}.} Each entry is a score's agreement with the output that a strict majority of readers chose, a tie counting as a miss~\citep{deutsch2023ties}. The pool rows cover the $173$, $163$, and $146$ pairs with a strict majority, drawn on the judge's earlier severities. The normalization-proof rows cover the $22$, $30$, and $30$ such pairs among them, a split made after the test. An asterisk marks a significant margin over normalized \cer{}. \Cref{fig:forest} gives every margin with its interval.}
\label{tab:decision}
\end{table}
}}
\@namedef{ax@float@tab_decision}{}
\@namedef{ax@float@fig_load}{\ax@pair[]{0.469}{\begin{figure}[t]
\centering
\includegraphics[max width=\linewidth]{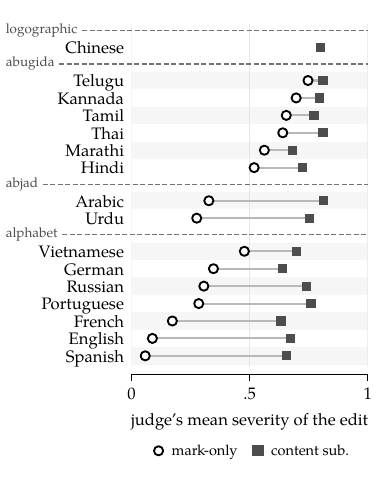}
\caption{\textbf{The cost of the smallest error varies by language.} The judge's mean severity of a mark-only edit (same base letter, different mark) and of a content substitution, grouped by writing system~\citep{daniels1996writing} and ordered by grapheme load (\cref{tab:grapheme-load}). An open circle is the mark-only edit and a filled square the content substitution.}
\label{fig:load}
\end{figure}
}{0.509}{\begin{figure}[!htb]
\centering
\includegraphics[max width=\linewidth]{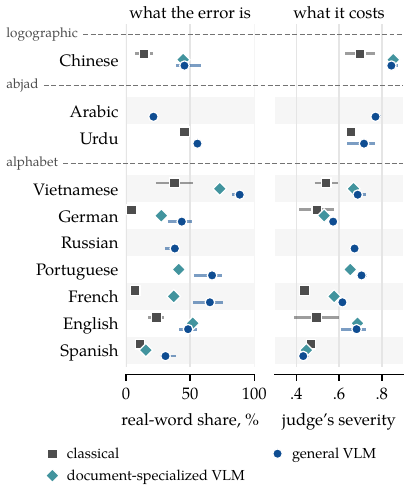}
\caption{\textbf{The form and cost of a letter error vary by system class.} A marker is the class mean and a bar spans its systems. The left panel is the share of a system's letter errors that form another real word, which needs no judge, and the right panel the judge's mean severity of those errors. The document-specialized VLM is Qianfan-OCR alone.}
\label{fig:system}
\end{figure}
}}
\@namedef{ax@float@fig_system}{}
\makeatother
\newcommand{\datasetcolwidth}{0.66\textwidth}
\AtBeginEnvironment{table}{\let\scriptsize\footnotesize\let\footnotesize\small}

\makeatletter\setlength{\@fptop}{0pt}\makeatother
\setcounter{topnumber}{4}\setcounter{bottomnumber}{3}\setcounter{totalnumber}{6}
\newcommand{\FloatBarrier}{}

\usepackage[capitalize]{cleveref}
\hypersetup{pdftitle={Equal CER, Unequal Damage: Severity-Weighted Character Error Rate for OCR},
  pdfauthor={Nguyen Nguyen, Yongyi Zang}}
\raggedbottom

\newcommand{\semcer}{\ensuremath{\text{sem\-CER}}}
\newcommand{\cer}{\ensuremath{\text{CER}}}
\newcommand{\cfcer}{normalized \cer{}}
\definecolor{editink}{HTML}{B64342}

\title{Equal \cer{}, Unequal Damage: Severity-Weighted Character Error Rate for OCR}

\author{Nguyen Nguyen \\
  University of Rochester \\
  \texttt{n.nguyen@rochester.edu} \And
  Yongyi Zang \\
  Independent \\
  \texttt{zyy0116@gmail.com}}

\begin{document}
\maketitle
\begin{abstract}
Character Error Rate (\cer{}), the standard metric for OCR and vision--language models (VLMs), charges a typo as much as a wrong name. We introduce \semcer{}, which keeps \cer{}'s alignment and denominator and charges each grapheme edit a severity from an LLM judge, with no word segmentation. Normalized \cer{} keeps the score of a whole transcription, where it tracks readers as well as \semcer{} or better, and \semcer{} breaks that score down into named edits. The charges track readers' ratings edit by edit in English, Vietnamese, and Chinese and put the edits readers rate most severe near the top of a sorted list. On pairs drawn where \cer{} and \semcer{} disagreed, readers side with \semcer{} more often than with normalized \cer{} in English and Chinese, and equally often in Vietnamese. Across $16$ languages, the judge charges a wrong diacritic or tone mark almost nothing in Spanish but nearly a wrong letter's severity in Telugu. Where we can count real words, a general VLM's wrong letter forms one far above chance while a classical pipeline's stays near it. We release the $61$ million judged edits, the aggregated reader ratings, and the distilled scorer at \url{https://github.com/nguyenx2t/semcer}.
\end{abstract}

\begin{center}
  \includegraphics[width=\textwidth]{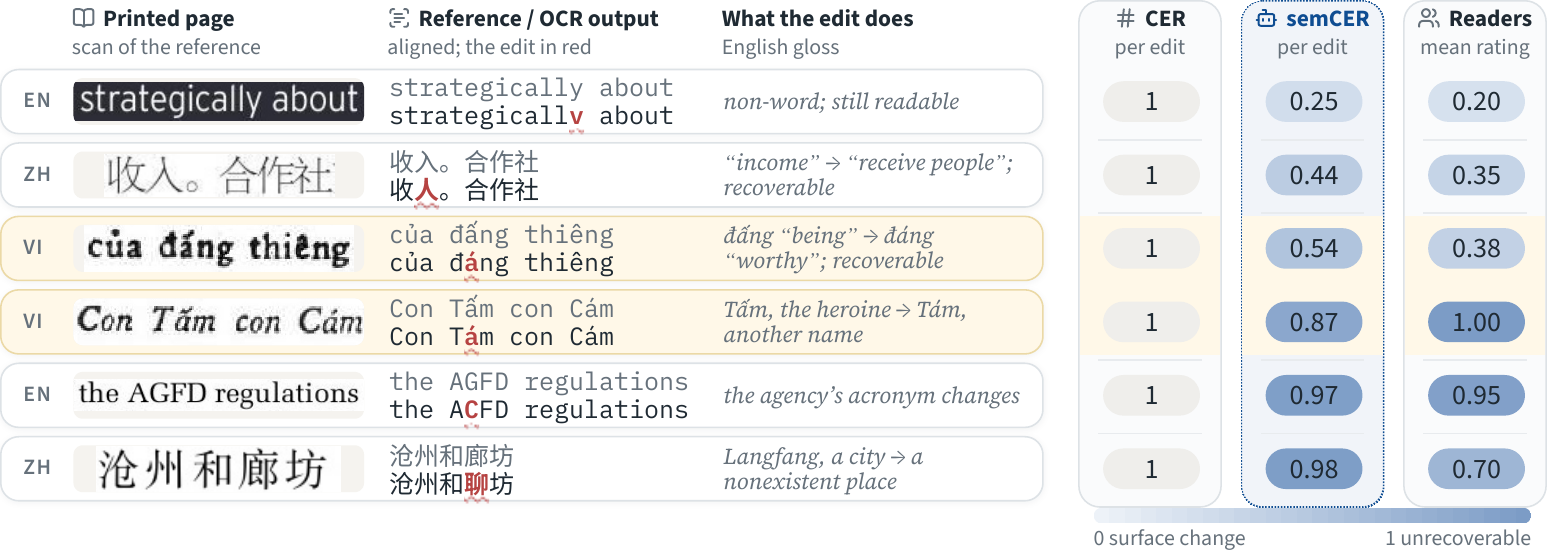}%

  \vspace{-8pt}%
  \captionof{figure}{\textbf{Equal edit counts can hide errors a reader cannot repair.} Six single-edit errors are sorted by \semcer{}'s charge; \cer{} charges each one equally. The highlighted rows share one substitution: \vn{đấng} (``being'') becomes the recoverable \vn{đáng} (``worthy''), whereas \vn{Tấm}, the heroine of \vn{\emph{Tấm Cám}}, becomes \vn{Tám}, another Vietnamese name.}%
  \label{fig:teaser}%
\end{center}
\section{Introduction}
\label{sec:intro}

Consider two optical character recognition (OCR) systems that each make one error on an English line (\cref{fig:teaser}). One writes \emph{strategically} as \emph{strategicallv}, and the other misreads the agency acronym \emph{AGFD} as \emph{ACFD}. A reader fixes the first without hesitation but cannot fix the second, which silently renames an organization. Yet Character Error Rate (\cer{}), the standard metric for OCR and vision--language models (VLMs) alike~\cite{omnidocbench2025,neudecker2021survey,beyene2026survey}, charges the two the same, because it counts edits and divides by the reference length, whatever each edit costs the reader.

Benchmarks and scoring toolkits soften the uniform charge by removing or folding whitespace, case, and punctuation before counting~\cite{omnidocbench2025,ccocr2024,jiwer}, but normalization still charges every surviving edit alike. In \cref{fig:teaser}, two Vietnamese lines contain the same substitution, \vn{ấ}$\to$\vn{á}. Readers rate the first recoverable from its sentence and the second unrecoverable because it changes the heroine's name, while normalization charges both the same. VLMs add a failure mode of their own, since they decode through a strong language prior and write fluent wrong words~\cite{gardella2026,karamolegkou2026reading,liang2026crutch,lee2026faithc4} that \cer{} charges no more than a typo.

Metrics that do weigh meaning either rate the whole output, as semantic similarity scores and most LLM judges do, or grade spans marked outside \cer{}'s alignment, as MQM does~\citep{freitag2021}. Neither stays on \cer{}'s scale. A rate that keeps \cer{}'s form and changes its charge raises three questions. (Q1) Which unit do we charge when a script puts no spaces between words? (Q2) Are a few severity classes fine enough to choose between two transcriptions that both read almost right? (Q3) Where does a severity still help once a benchmark has normalized cosmetic edits away?

In this paper, we introduce \semcer{}, which answers Q1 by design. It charges the grapheme clusters that \cer{}'s alignment already produces and keeps \cer{}'s denominator. One definition therefore applies to alphabets, abugidas, abjads, and unspaced scripts without word segmentation, and charging every edit one gives back \cer{} exactly. For Q2, a judge rates each edit on a five-level rubric, and \semcer{} reads the charge from the judge's probabilities rather than its most likely level, so two edits at the same level can still differ. Whether that finer charge agrees with readers and chooses between two almost-right transcriptions better than a three-class charge does (Q2), and whether it adds to normalization (Q3), are empirical questions. We answer them by testing \semcer{} against readers, a three-class charge, and normalized \cer{}. The answers make \semcer{} a tool for finding and grading errors, while normalized \cer{} remains the score for a whole transcription. We make four contributions.
\begin{itemize}[leftmargin=*,nosep,itemsep=2pt]
\item \textbf{\semcer{} orders the errors to check.} Sorted by charge, the edits readers rate most severe come near the top, an order \cer{} cannot give. The charge tracks readers' ratings beyond what the edit's type predicts, and in English and Chinese beyond whether normalization would remove the edit (\cref{sec:grade}).
\item \textbf{Readers mostly side with \semcer{} where it disagrees with \cer{}.} On pairs drawn where \cer{} and \semcer{} ranked two transcriptions in opposite order, readers of English and Chinese side with \semcer{} on $71$ and $78\%$ of pairs. They side with normalized \cer{}, the stronger baseline, on $58$ and $60\%$, and Vietnamese readers side with the two equally often. On the same pairs, readers of all three languages side with \semcer{} significantly more often than with a three-class charge (\cref{sec:beyond}).
\item \textbf{An equal edit count hides unequal damage, set by the language and by the system.} The judge's charge for an edit varies across $16$ languages with the price of the smallest edit, which costs almost nothing in Spanish and nearly a wrong letter's severity in Telugu. Where a general VLM and EasyOCR tie on edit count, the VLM's edits cost more in Chinese and Vietnamese and less in English, French, and Spanish. The split follows the price of the smallest edit, and at least three of the four VLMs show it in each language. The VLM's letter errors cost more than a classical pipeline's in every language where we can measure both, except Spanish, and they form real words where the pipeline's form non-words, so a word-list check misses them (\cref{sec:corpus}).
\item \textbf{A distilled scorer with fixed weights reproduces the judge.} A $0.8$B student trained on the judge's distributions, with no human rating, comes within $0.03$ of the judge's agreement with readers and orders nine in ten clear edit pairs as the judge does (\cref{sec:student}). We release it with the $61$ million judged edits and the aggregated per-item reader ratings.
\end{itemize}

\section{Related Work}
\label{sec:related}

\paragraph{Downstream harm.} OCR errors matter unequally downstream. They corrupt entities and
numbers in retrieval and question
answering~\citep{vanstrien2020,hamdi2023ner,jaud2025beyond,ohrbench2025}, and a low \cer{} can hide
decision-critical errors~\citep{fincriticaled2025}.

\paragraph{Error rates with a graded charge.} Speech recognition has several error rates that keep
the word error rate's form and replace its unit charge. EmbER discounts substitutions with similar word embeddings~\citep{baneras2022ember}, Semantic-WER weights entities and sentiment words~\citep{roy2021semanticwer}, and SWWER weights words by BERT-derived importance~\citep{zhang2024swwer}. Three others grade each error by its effect, minED~\citep{baneras2026mined}, clinicians' impact labels~\citep{ellis2025werunaware}, and the NER accuracy rate for live subtitling~\citep{romerofresco2015ner}.

\paragraph{Embedding similarity as semantic severity.} BERTScore, BLEURT, SemDist, and H\_eval
compare the whole reference and hypothesis through embeddings or learned
scores~\citep{bertscore2020,bleurt2020,semdist2021,heval2023}, and aligned variants score token or
segment pairs~\citep{rugayan2022asd,sasindran2024semascore}. They charge a similarity on a scale
unrelated to \cer{}'s, and BERTScore responds weakly to small errors in lexically close candidates,
including a changed figure~\citep{hanna2021bertscore,huang2025finnue}, which are the errors OCR makes.

\paragraph{Model-judged severity on spans.} MQM annotates error spans by type and weights them by
severity~\citep{lommel2014mqm,freitag2021}, and GEMBA-MQM, AutoMQM, xCOMET, and TIGERScore have a
model mark and grade the spans~\citep{gembamqm2023,automqm2023,xcomet2024,jiang2024tigerscore}. In
grammatical error correction, CLEME2.0 also weights individual edits with a language model~\citep{cleme2025}. It aligns the
erroneous source, the system's correction, and the reference corrections into word chunks and scores
each chunk as a hit, a wrong, a missing, or an unnecessary correction. It then weights a chunk by embedding
similarity or by an LLM's rating of how important the correction is and combines the four scores
with fitted coefficients. JELV and LLM judges likewise rate a correction from its marked
edits~\citep{zhan2025jelv,kobayashi2024llmgec}. \semcer{} has only a reference and no erroneous source text, so it scores
the recoverability of each error in the transcription, where CLEME2.0 scores the importance of each correction.
It charges the graphemes of a fixed unit-cost alignment, which needs no word segmentation, and keeps
\cer{}'s denominator, so charging every edit one gives \cer{} back.

\paragraph{LLM judges for document transcription.} LLM judges rate extracted
tables~\citep{horn2026table}, penalize handwritten-math transcriptions that silently correct the
writer~\citep{seong2026pink}, and sort the errors of a text diff into minor and
major~\citep{gutteridge2026judge}. Each scores a whole output or counts errors by class.

\paragraph{Normalization and cost tables.} OCR benchmarks transform text differently before
scoring~\citep{omnidocbench2025,ccocr2024,rice1996isri}, and in speech the choice is known to distort
evaluation, since Whisper's normalizer strips Indic vowel signs and flatters the
model~\citep{manohar2024normalization,thennal2025cer}. A second line of work replaces the uniform
Levenshtein costs~\citep{levenshtein1966} with learned or tuned ones, either a weight per operation or
character pair that is fitted once and reused~\citep{ristad1998,oncina2006,eed2019} or operation
probabilities from learned encodings of both strings~\citep{libovicky2022}. Normalization rewrites the
strings before alignment, and cost tables let the costs choose the alignment. \semcer{} instead fixes a
unit-cost alignment first and charges its edits afterward.

\paragraph{Output-level OCR metrics.} Other OCR metrics decompose page-level
error by pipeline stage~\citep{cev2026}, ignore reading order~\citep{clausner2020flexible}, or score
reconstruction fidelity against the page image~\citep{vrtf2026}. SCORE, for generative parsing, adjusts the edit distance for representational variation and counts hallucinated and omitted tokens~\citep{score2025}. It normalizes the variation away before counting, while \semcer{} charges
what survives. Unit-test benchmarks replace the edit count with binary checks per
page~\citep{olmocr2}.

\section{The \semcer{} Metric}
\label{sec:metric}

\semcer{} has two parts, a definition that puts a severity in place of \cer{}'s unit charge and a judge that supplies the severity.

\subsection{Definition}
\label{sec:metric-def}
\semcer{} replaces each edit's unit charge in \cer{} with a severity, so the first choice is what counts as one edit. Like \cer{}, it charges characters rather than words, because a script with no spaces between words yields a word error rate only after a segmentation step whose mistakes the score inherits~\citep{thennal2025cer} (Q1). We treat a character in both scores as an extended grapheme cluster, following Unicode Standard Annex \#29~\citep{uax29} under Normalization Form C, so a base letter with its accents or tone marks is one unit, as grapheme-level OCR metrics recommend~\citep{nisfer2026graphemekit}. Published \cer{} implementations often count code points instead~\citep{jiwer}, which splits a syllable of an abugida into its consonant and vowel signs and counts an accented letter as one unit or several depending on how the text was encoded. Extended grapheme clusters approximate the characters a reader perceives and give the same units whatever the encoding form, so the judge rates one kind of unit in every language. We compute every \cer{} in this work over the same grapheme clusters, case- and whitespace-sensitive, and call it \emph{unnormalized} \cer{}.

Let the reference and output be grapheme sequences $r=(r_1,\dots,r_m)$ and $h=(h_1,\dots,h_n)$. An optimal Levenshtein alignment~\citep{levenshtein1966,wagner1974} yields the edit sequence $\mathcal{E}(r,h)=(e_1,\dots,e_q)$ of substitutions, deletions, and insertions. We break ties among minimal alignments toward substitutions, then deletions, so the judge sees a misread glyph as one edit, and \cer{} is unchanged. For $m>0$, \cer{} charges each edit one and divides by the reference length,
\begin{equation}
  \cer(r,h)=\frac{1}{m}\sum_{i=1}^{q} 1=\frac{q}{m}.
  \label{eq:cer}
\end{equation}
\semcer{} keeps this form and replaces the unit charge by a severity $s_i=s(e_i;r,h)\in[0,1]$, the judged cost of edit $e_i$ read in its line,
\begin{equation}
  \semcer(r,h)=\frac{1}{m}\sum_{i=1}^{q} s_i,
  \label{eq:semcer}
\end{equation}
where zero marks a surface change, a low value a recoverable edit, and one an unrecoverable loss of content.

\semcer{} therefore discounts the edits the judge rates recoverable. With every $s_i=1$ it is \cer{} itself, and it never exceeds \cer{}, since no severity exceeds one.

\begin{figure}[t]
\centering
\input{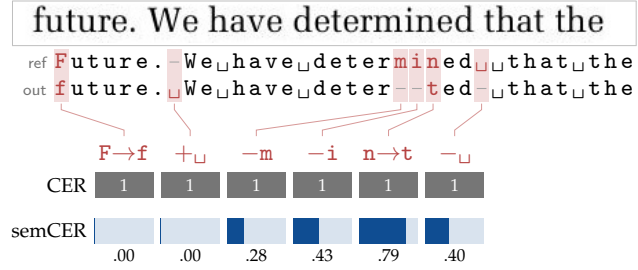}
\caption{\textbf{The metric charges one transcription line edit by edit.} A general VLM transcribes an OmniDocBench line, shown as the crop above and the aligned reference and output below. Each edit is shaded, $+$ marks an insertion, $-$ a deletion, and a dash an empty side. Six edits give \cer{} $0.171$ over $35$ graphemes, while their severities sum to $1.90$ and give \semcer{} $0.054$.}
\label{fig:priced}
\end{figure}

\Cref{fig:priced} scores one line. The output's lowercase \emph{f} follows the scan where the reference capitalizes it, so it costs nothing, and so does an added space. One wrong word, which the alignment splits into three edits, carries most of the charge, and a dropped space the rest.

\subsection{Severity from an LLM judge}
\label{sec:metric-judge}
Each edit's severity comes from a large language model judge, DeepSeek-V4-Flash Preview~\citep{deepseekv4}. It reads the full reference and output with the target edit marked in context (\cref{app:prompt}). It then rates how far the edit stops a reader from noticing and repairing it on the ordinal rubric of \cref{tab:rubric}, from $0$ for a surface change to $4$ for an error that leaves no signal. We fit no parameter to human ratings.

\begin{table}[t]
\centering\footnotesize
\setlength{\tabcolsep}{3pt}
\begin{tabular}{@{}ccl@{}}
\toprule
Level & Rating & Description \\
\midrule
$0$ & $0$    & Cosmetic, fully recoverable \\
$1$ & $0.25$ & Non-word or obvious error, context recovers \\
$2$ & $0.5$  & Valid alternative, reader may notice \\
$3$ & $0.75$ & Recovery fails, but error is detectable \\
$4$ & $1$    & Reads perfectly, is false, no signal \\
\bottomrule
\end{tabular}
\caption{\textbf{The rubric's five levels.} The judge returns a level, and its rating $\rho_k=k/4$ is the scale of the severity and of the readers' ratings.}
\label{tab:rubric}
\end{table}

The severity separates edits that share a rubric level, because we read the judge's probabilities over the rubric rather than its most likely level, the mode. G-Eval~\citep{geval2023} and LLM-as-a-Verifier~\citep{kwok2026verifier} read probabilities the same way, and judgment-distribution studies recommend it~\citep{wang2025judgmentdist}. Scores that take only the most likely level tie often, and their rankings do not reproduce~\citep{kim2026tablejudge}. Let $p_k$ be the probability the judge assigns to digit $k\in\{0,\dots,4\}$ at its answer position, renormalized over the five digits,
\begin{equation}
  \tilde p_k=\frac{p_k}{\sum_{j=0}^{4} p_j}.
\end{equation}
Each level $k$ carries the rating $\rho_k=k/4$ of \cref{tab:rubric}, and the severity is the expected rating under $\tilde p$, which treats the rubric's steps as equally spaced,
\begin{equation}
  s_i=\mathbb{E}_{k\sim\tilde p}\,[\rho_k]=\sum_{k=0}^{4}\rho_k\,\tilde p_k.
  \label{eq:severity}
\end{equation}
When the judge returns no digit, or puts less than half of its probability on the five digits, we ask once more and otherwise charge the edit one (\cref{app:prompt}).

\section{Experimental Setup}
\label{sec:setup}

\begin{figure*}[t]
\centering
\includegraphics[max width=\linewidth]{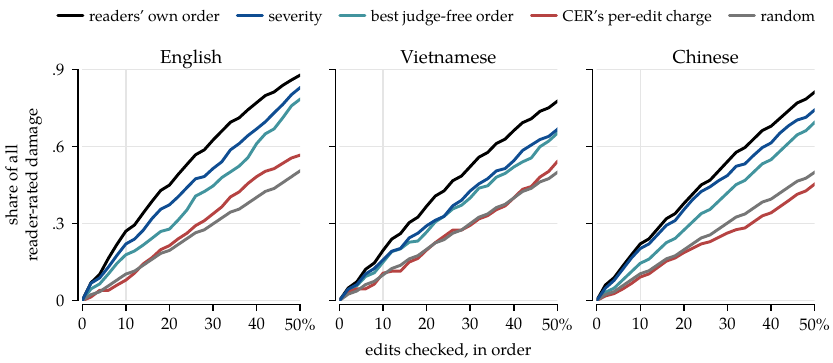}
\caption{\textbf{Sorted by severity, the edits readers rate most severe come first.} Each panel gives the share of all reader-rated damage caught after checking a fraction of the single-edit pool in each order, over the first half of the list. The readers' own order is the ceiling, a random order the diagonal, and the rule marks the $10\%$ cut.}
\label{fig:review}
\end{figure*}

\paragraph{Corpus.} We run the judge on the outputs of seven systems on $25$ public datasets in $16$ languages, $61$ million edits in all, scored line by line after the filtering in \cref{app:corpus}. The systems are of three kinds, two classical pipelines, a document-specialized VLM, and four general VLMs (\cref{tab:systems}).

\paragraph{Human annotation.} Native or fluent readers of English, Vietnamese, and Chinese annotated three pools, sized in \cref{tab:reliability} (\cref{app:annotation}). The \emph{single-edit} pool holds about $85$ edits per language, each the only edit in its line, and tests whether the judge charges an edit as a reader would. We drew it evenly from three bands of the judge's severity, low, intermediate, and high, so the pool spans the scale. The \emph{pair-level} pool, named for its reference and transcription pairs, holds about $180$ whole transcriptions per language and tests the summed score. Readers rate both on the rubric's five levels. The \emph{preference} pool holds about $175$ pairs of transcriptions of one line per language, drawn where \cer{} and \semcer{}, on the judge's earlier severities, ranked them in opposite order, and readers choose the better OCR of the reference.

\paragraph{Baselines.}\label{sec:baselines} We compare the severity with three charges that use no judge, a table built from the judge's corpus, and two coarser readouts of the same judge. The judge-free charges are \cer{}'s own per-edit charge, the \emph{filter label}, and the \emph{real-word label}. The filter label records whether an edit survives the character filter that OCR benchmarks apply before scoring, which keeps letters and digits and drops whitespace, punctuation, and symbols~\citep{omnidocbench2025}, and the real-word label whether the edit forms a real word. The \emph{type table} charges an edit the corpus-mean severity of its type and reads the judge's corpus but not the line. The coarser readouts are the \emph{class charge}, the judge's most likely level charged on three fixed class weights, and the \emph{whole-line judgment}, one rating of the whole line. Normalized \cer{} applies the filter and folds case before alignment.

\paragraph{Statistics.} Intervals are $95\%$ percentile bootstrap intervals from $10{,}000$ draws, resampling items within band for the single-edit partial correlations and source documents elsewhere. The single-edit partial correlations control for the sampling band, since we drew that pool by the judge's severity. A partial whose interval excludes zero therefore shows agreement within a band, which the draw alone cannot produce. We take a comparison's $p$-value as twice the smaller bootstrap tail probability about zero and call a difference significant at $\alpha=0.05$ after Holm correction~\citep{holm1979} within the families that \cref{app:stats} names.

\section{Per-Edit Agreement with Readers}
\label{sec:grade}

A user who checks a transcription wants to know which errors to look at first, and a score alone does not say. \semcer{} also orders the edits behind its score, so we first test that order against readers and then ask what the severity adds to the simpler charges an order could use.

\subsection{Which errors to check first}
\label{sec:grade-order}

Sorted by \semcer{}'s charge, the edits readers rate most severe come near the top. \Cref{fig:teaser} is such a list. \cer{} charges its six edits the same and so cannot order them, while the three highest severities go to the three edits that readers rate most severe.\label{sec:grade-review}

On the single-edit pool each edit sits in its own line, so an order of the pool runs across lines. \Cref{fig:review} follows each order through the pool and shows the share of all reader-rated damage, the sum of readers' mean ratings, caught after checking a fraction of the edits. The \emph{top-of-list share} summarizes that curve over cuts at $10$ to $50\%$ of the list, scaled so that a random order scores $0$ and the readers' own order $1$ (\cref{app:review}).

Sorted by severity, the list scores $0.58$ to $0.81$ across the three languages, and its top $10\%$ catches $81$ to $92\%$ of the damage that the readers' own top $10\%$ catches (\cref{tab:grade-full}). Sorted by \cer{}'s per-edit charge, $1/m$, which varies only with the line's length, the list scores about zero, the random baseline. Its top $10\%$ catches $29$ to $55\%$, a gap that is significant in all three languages (\cref{tab:grade-full}). The best order that uses no judge, chosen among four sorted by the filter label, the real-word label, \cer{}'s per-edit charge, or all three (\cref{app:baselines}), lands about halfway. Severity's lead over it on the share is significant in Chinese alone (\cref{tab:grade-full}).

\label{sec:grade-edit}\label{sec:grade-bit}The order holds because the charges match what readers make of each edit. On the single-edit pool, the severity correlates with readers' mean rating in all three languages (\cref{tab:grade-full}).

\float{fig_grade}

\subsection{What the severity adds}
\label{sec:grade-adds}

\paragraph{Beyond the edit's type.}\label{sec:grade-table} A charge that ignores the line falls short, because the line changes what an edit costs. The type table looks an edit up by its operation and its reference and output graphemes and charges it the corpus-mean severity of that type, computed without the single-edit pool. For \emph{strategicallv} in \cref{fig:teaser}, it charges $0.59$, the mean for a \emph{y}$\to$\emph{v} swap. The judge, which reads the line, charges
$0.25$, and readers give $0.20$. Controlling for the table, the severity still tracks readers in all
three languages (\cref{fig:grade}, type-table row).

\paragraph{Beyond normalization.}\label{sec:grade-flag} In English and Chinese, the severity also adds to the normalization that benchmarks already apply (Q3), because it orders the edits normalization keeps. Pooled over the three bands, the filter label correlates with readers about as well as the severity does (\cref{tab:grade-full}). Within a band, the severity still adds agreement once we control for the label alone or for the label and the type table together. Its intervals lie above zero in English and Chinese (\cref{fig:grade}, bottom two rows).

\paragraph{Beyond the most likely level.} Reading the judge's full distribution rather than its most likely level adds agreement with readers in Vietnamese and Chinese (\cref{tab:grade-full}, severity given the mode).

\section{Reader Preferences Where \cer{} and \semcer{} Disagree}
\label{sec:beyond}

On whole transcriptions, normalized \cer{} tracks readers as well as \semcer{} or better. Its Spearman's $\rho$ is $0.93$, $0.90$, and $0.76$ against \semcer{}'s $0.89$, $0.87$, and $0.76$ (\cref{tab:family}).\label{sec:family-lever} \semcer{}'s value beside \cer{} therefore lies where the two scores rank a pair of transcriptions of one line in opposite order. We ask how often readers side with \semcer{} there, and whether it adds to what normalization already catches.

\float{tab_decision}

\paragraph{Where the scores disagree.}\label{sec:beyond-decision} We drew the preference pool from such disagreements on the judge's earlier severities, one pair per source line, and readers chose the better transcription in each. A score agrees with a pair when it picks the output a strict majority of readers chose, a tie counting as a miss. When the pool was drawn, unnormalized \cer{} picked the other output on every pair, so its agreement was one minus \semcer{}'s, and the baseline is therefore normalized \cer{}, which can side with either. On the whole pool, readers of English and Chinese side with \semcer{} on more than seven pairs in ten, against about six in ten for normalized \cer{}. Vietnamese readers side with the two scores equally often, at about eight in ten (\cref{tab:decision}, top).

\paragraph{Near-miss pairs.} Two nested subsets sharpen the test. In a \emph{near-miss pair}, the output that \cer{} rates worse is mostly right and has an error that normalization would keep (\cref{app:decision-test}). Before the test, we fixed the criterion that \semcer{} lead normalized \cer{} significantly in at least two languages on these pairs. It does, in English and Chinese (\cref{tab:decision-detail}).

\paragraph{Normalization-proof pairs.} A near-miss pair is \emph{normalization-proof} when normalized \cer{} picks the same output as \cer{}, a split that uses no severity. \semcer{} matters most here, where normalization leaves \cer{}'s choice in place (Q3). On these pairs, normalized \cer{} almost never agrees with English readers and agrees with Chinese readers on under half the pairs, while \semcer{} agrees with most. In Vietnamese, both scores agree with readers on about half the pairs (\cref{tab:decision}, bottom). The English and Chinese margins on these pairs also hold when scored with the judge's earlier severities, with the distilled scorer, and with the second judge of \cref{app:second-judge} (\cref{fig:forest}).

\paragraph{Coarser charges.} Coarser readouts of the same judge agree with readers less often than \semcer{} (Q2). The class charge keeps the judge's labels and charges them on the three fixed weights that severity-aware WER assigns its OK, minor, and critical classes~\citep{wang2024swer}. On the whole pool, it trails \semcer{} significantly in every language, as the type table does (\cref{fig:forest}). One rating of the whole transcription is coarser still. If it sufficed, the judge could rate the line once with a severity $\bar{s}$, and \cref{eq:semcer} would reduce to $\cer{}\cdot\bar{s}$. This whole-line judgment trails \semcer{} significantly in English and Vietnamese (\cref{app:holistic}).

\section{Severity by Language and by System}
\label{sec:corpus}

An equal edit count hides unequal damage. When two transcriptions of a line make the same number of edits, \cer{} calls it a tie, and \semcer{} breaks the tie by the edits' severities. Where a general VLM and EasyOCR tie on one to five edits, the VLM's edits cost more in Chinese and Vietnamese and less in English, French, and Spanish (\cref{tab:equal-cost-app}). At least three of the four VLMs agree in each language, German shows no difference, and the split follows the languages' grapheme load (\cref{tab:grapheme-load}). The gap is significant for $8$ of the $24$ pairings after correction over all $24$, short of the criterion we fixed, a VLM charge above EasyOCR's for every VLM (\cref{app:equal-cost}). A count sees neither what the language charges for its smallest edit nor what kind of error the system makes.

\float{fig_load}

\paragraph{Language.} The judge charges a wrong letter between $0.63$ and $0.81$ across scripts, but a wrong mark anywhere from nearly nothing in Spanish to almost a wrong letter's severity in Telugu. We measure the cost of the smallest error with the \emph{grapheme load}, the judge's mean severity of the smallest word-changing edit that systems make. Where the script has marks, that edit changes only a mark, a diacritic, vowel sign, or tone mark on the same base letter. In Chinese it is a content substitution, one base letter for another (\cref{fig:load}). The load depends on the marks a script writes more than on the script's type, so Vietnamese, an alphabet with
tone marks, sits near Hindi. The
mean severity per edit follows the load, with a rank correlation of $0.75$ across the $15$ languages with marks, from $0.26$ in English to $0.74$ in Telugu
(\cref{tab:severity-per-edit}).

\paragraph{System.} A general VLM's letter error forms a real word far more often than a random letter error of the same shape would, as prior work expects of a system with a stronger
language prior~\citep{wan2020vocabulary,karamolegkou2026reading,liang2026crutch,he2025seeing,gardella2026,lee2026faithc4}.
EasyOCR reads \emph{by} as \emph{bv}, while a VLM writes \emph{has} for \emph{had}. The share of letter errors that form a real word needs no judge to count. In German, French,
Spanish, and Vietnamese it rises from the classical pipelines through the document-specialized VLM to the general ones. In the ten languages where we can count real words, general VLM errors form them more often than random letter errors in every language and domain but one, printed, handwritten, and scene text alike. Classical errors stay within $0.10$ of the random rate. In each of the eight language and domain sets where both make enough letter errors to compare, the VLM's margin over random errors is the larger (\cref{app:behavior}).

The judge charges these errors more. A real-word error passes a word-list check~\citep{kukich1992}, yet the judge
charges a general VLM's letter error more than a classical pipeline's in every language where we can
measure both, except Spanish (\cref{fig:system}).

\section{A Distilled Scorer with Fixed Weights}
\label{sec:student}

A small model with fixed weights approximates the judge, so anyone can compute \semcer{} locally. Scoring edits through an API costs money at scale and ties the scores to whichever model the provider serves. We distill the judge into a student, Qwen3.5-0.8B-Base~\citep{qwen35blog}, which we fine-tune to read the pair with the
edit marked and predict the judge's distribution over the rubric levels, read out as in
\cref{eq:severity}. We hold out every rated-pool line under every system and use no human rating for
training or model selection (\cref{app:student}). The student ships as a Python package, installed with \texttt{pip install semcer}.

The student nearly matches the judge's agreement with readers. It correlates with their ratings at $0.79$, $0.63$, and $0.80$ in English, Vietnamese, and Chinese, against the judge's $0.81$, $0.66$, and $0.82$. On pairs of edits whose judge severities differ by at least half a rubric step, it orders nine in ten as the judge does, and its severity differs from the judge's by a mean absolute $0.04$ per edit (\cref{app:student}).

\section{Conclusion}
\label{sec:conclusion}

\cer{} counts edits, and normalization decides which edits to count, but neither says what an edit costs the reader. We introduced \semcer{}, which charges each grapheme edit a judged severity, without word segmentation, and keeps \cer{}'s alignment and denominator. The two scores divide the work. Normalized \cer{} tracks readers' ratings of whole transcriptions as well as \semcer{} or better, so it remains the score for a transcription. An equal edit count is a tie for \cer{}, and \semcer{} breaks it with charges that follow the language's smallest edit and the system's kind of error. Where the two scores order a pair of transcriptions the other way, readers of English and Chinese mostly side with \semcer{}. They do so most of all where normalization leaves \cer{}'s choice in place, and readers of all three languages side with \semcer{} over a three-class charge. The score is a sum over the aligned edits, so each decision between two transcriptions breaks down into named edits and their charges. Those charges track readers' ratings edit by edit in three languages and put the errors readers rate most severe near the top of a sorted list. We therefore recommend normalized \cer{} for scoring a transcription. Beside it, \semcer{}, computed locally with the distilled scorer, chooses between transcriptions where \cer{} ties or the two scores disagree, and names the edits behind the choice.

\section{Limitations}
\label{sec:limitations}

\paragraph{Languages.} Readers validated the severity in three languages. In the other thirteen, the
grapheme load comes from the judge alone, and the judge's severities may shift with a language's
resource level~\citep{joshi2020state}. The lowest-resource languages in our corpus all use non-Latin scripts, so we cannot
separate resource level from script. We have not tested whether graphemes are the right unit in Thai, which writes no word boundaries, or in the abjads, which leave vowels unwritten.

\paragraph{Construct.} Readers rated on the rubric's levels, and preference readers chose after
guidelines that teach the same construct. The human studies therefore test whether the judge applies
the rubric as readers do and leave open whether recoverability is the right construct. The rubric's levels
depend on whether an output is a valid word, so part of the higher charge the judge places on general VLM letter
errors in \cref{sec:corpus}, which more often form real words, comes from the rubric itself. We leave \semcer{}'s value for downstream
tasks to future work.

\paragraph{Judge version.} The provider replaced the judge's checkpoint during this work, so a
severity is comparable only within one checkpoint. Corpus severities come from the checkpoint before
the replacement. In the rated pools only the edits the judge first left undecided were re-judged on the one after, and the English and Chinese
margins on the preference pool and its normalization-proof pairs hold under both (\cref{app:judge-version}). The local student has fixed weights, so it avoids this
boundary at the cost of approximating the judge.

\paragraph{Rating.} Readers rated each single edit as the only edit in its line and rated whole transcriptions without marked
edits, so no reader rated an edit's severity inside a line with other damage. The judge charges each
grapheme edit separately, so a wrong word that the alignment splits into several edits pays once per
edit (\cref{fig:priced}).

\paragraph{Risks.} A low \semcer{} says that a reader could recover the errors, and a transcription with a low score
still needs checking before use. A judge that undercharges an error class would hide that class.

\section*{Use of AI Assistants}
Large language models are part of the method. They judge each edit's severity, a second one re-judges every rated transcription, and the distilled scorer is a fine-tuned language model. For
this manuscript, we used the same class of tools for copy editing (grammar, clarity, and formatting)
and for drafting support. We also used language models and agents for literature search, code
writing, debugging, and informal discussion and brainstorming. The authors are solely responsible for
the scientific content, study design, experiments, and conclusions.

\clearpage
\appendix
\crefalias{section}{appendix}
\crefalias{subsection}{appendix}
\crefname{appendix}{Appendix}{Appendices}
\Crefname{appendix}{Appendix}{Appendices}
\section{Setup Details}

\subsection{Judge configuration}
\label{app:prompt}

DeepSeek-V4-Flash Preview scores each marked edit at temperature $1$, answering directly without a reasoning trace. It sees the full reference and output, then the edit in brackets, and returns one of the five rubric levels of \cref{tab:rubric}.

\paragraph{Prompt.} The system message is the text below, the same for every per-edit severity in this paper, while the whole-line judgment of \cref{app:baselines} uses its own prompt. The user message gives the reference and output in full as \texttt{REFERENCE:} and \texttt{HYPOTHESIS:} blocks, then the edit block in the system message's notation with ten graphemes of context on each side and an ellipsis beyond, then the line \texttt{Output exactly one digit (0, 1, 2, 3, or 4):}.

\begin{Verbatim}[fontsize=\scriptsize,breaklines=true,breakindent=0pt,breaksymbolleft={},frame=lines,framerule=0.4pt,framesep=3pt]
You are an OCR quality evaluator. For each request, rate how much a single grapheme-level edit damages semantic meaning, given the surrounding context.

TASK: You are given an edit operation and two views of the same sentence — REF with the edit position marked in brackets, and HYP with the edit position marked in brackets. Score the damage on integer scale 0-4 based on the rubric.

1. Read REF and HYP to establish sentence-level context.
2. Identify the operation type and locate the edit at the bracket position.
3. Determine if the hypothesis token is a valid word/morpheme in the language.
4. Determine if the sentence-level context signals an error and if the original is recoverable.
5. Apply SEVERITY LEVELS strictly. Do not score on intuition.
6. Output exactly one digit: 0, 1, 2, 3, or 4. No reasoning text.

LANGUAGES: Any natural language. Apply identical criteria regardless of script.

EDIT NOTATION:
- SUB: REF contains [X], HYP contains [Y]
- DEL: REF contains [X], HYP contains []
- INS: REF contains [], HYP contains [Y]

SEVERITY LEVELS:

0 — Surface only. Case, punctuation, whitespace, or glyph variant. Meaning fully preserved; change would not be noticed in normal reading.

1 — Minor. Hypothesis is a non-word but sentence-level context uniquely determines the original. Reader recovers with negligible effort.

2 — Moderate. Hypothesis is a valid word but contextually plausible-but-odd in local syntax. Reader may parse at face value or sense an error; recovery requires noticing the mismatch.

3 — Severe with partial recovery. Result signals an error (grammatical anomaly, format violation, collocational mismatch) but the original cannot be uniquely recovered from candidate set.

4 — Total. Hypothesis is fully valid, fits surrounding context, gives no signal of error, original unrecoverable from this edit alone. Magnitude doesn't matter; what matters is recoverability.

ANCHOR (Score 4):
Operation: SUB
REF: The team meeting will be held at [3]pm tomorrow in the conference room.
HYP: The team meeting will be held at [8]pm tomorrow in the conference room.
Why 4: Sentence context "team meeting ... conference room" doesn't disambiguate the time. "8pm" is fully valid. Reader has no path to recover "3pm". Information destroyed.

EXAMPLES:

Operation: SUB
REF: I will meet you at 3p[m] tomorrow at the cafe.
HYP: I will meet you at 3p[M] tomorrow at the cafe.
Score: 0

Operation: SUB
REF: The patient is im[p]roving rapidly after surgery.
HYP: The patient is im[b]roving rapidly after surgery.
Score: 1

Operation: SUB
REF: Please turn l[e]ft at the next corner near the bank.
HYP: Please turn l[i]ft at the next corner near the bank.
Score: 2

Operation: SUB
REF: She lives on the [1]3th floor of the building.
HYP: She lives on the [2]3th floor of the building.
Score: 3

Operation: SUB
REF: She was born in 19[9]9 in a small village.
HYP: She was born in 19[8]9 in a small village.
Score: 4

OUTPUT FORMAT: Exactly one digit: 0, 1, 2, 3, or 4. Nothing else.
\end{Verbatim}

\paragraph{Undecided edits.} When the judge returns no digit or puts less than half of its probability on the five digits, we ask it once more. If it fails again, we charge the edit one, the highest severity, the conservative choice for an edit the judge cannot decide. This imputed charge affects $0.8\%$ of corpus edits, between $1.6$ and $4.3\%$ in the three rated languages and under $0.01\%$ elsewhere. It enters only the corpus-level \semcer{} scores. The analyses of \cref{sec:corpus} use only transcriptions the judge scored completely, and we re-judged every undecided edit in the rated pools, so no rated edit carries an imputed charge (\cref{app:judge-version}).

\subsection{Corpus, systems, and compute}
\label{app:corpus}

\paragraph{Corpus.} We drop non-transcription tasks, references shorter than $20$ or longer than $4{,}000$ code points, records with only numbers or bare URLs, and formula-dense text. A reference counts as formula-dense when it is under $200$ code points with three or more \LaTeX{} commands, or under $100$ with more than a fifth of its characters mathematical symbols. We then keep at most $15{,}000$ lines per dataset and language for handwriting and printed documents and $3{,}000$ lines per language for scene text, pooled across datasets (\cref{tab:corpus-summary}).

\paragraph{Systems.} Qianfan-OCR and the general VLMs (\cref{tab:systems}) receive the prompt \texttt{Read all the text visible in this image. Output only the text, nothing else.} EasyOCR, PP-OCRv5, and Qianfan-OCR ran on one RTX 3090, and the four general VLMs took about $40$ H100-hours in all under vLLM~\citep{kwon2023vllm}. Judging the corpus cost about \$2{,}000, and the distilled scorer trained for about $30$ hours on one A100.

\paragraph{Release.} We use each dataset for research under its terms. We release severities and ratings under CC BY 4.0, with source text only where its license permits redistribution, and the scorer's code and weights under Apache-2.0. The other sources carry labels only, keyed by line identifier to match the original datasets.

\begin{table*}[!htb]
\centering\footnotesize
\setlength{\tabcolsep}{5pt}
\providecommand{\datasetcolwidth}{0.75\textwidth}
\begin{tabular}{@{}lp{\datasetcolwidth}r@{}}
\toprule
Type & Datasets & Lines \\
\midrule
Handwriting & IAM~\citep{marti2002iam}, RIMES~\citep{grosicki2011rimes}, German Handwriting~\citep{germanhw}, CASIA-HWDB~\citep{liu2011casia}, KHATT~\citep{mahmoud2014khatt}, UNHD~\citep{ahmed2019unhd}, VNOnDB~\citep{nguyen2018vnondb}, Thai handwriting~\citep{thaihw}, BRESSAY~\citep{neto2024bressay}, School Notebooks~\citep{schoolnotebooksru}, ICDAR 2025 IHDR~\citep{ihdr2025} & 104{,}897 \\
Scene text & HierText~\citep{long2022hiertext}, LSVT~\citep{sun2019lsvt}, RCTW-17~\citep{shi2017rctw}, COCO-Text~\citep{veit2016cocotext}, TextOCR~\citep{singh2021textocr}, CTW~\citep{yuan2019ctw}, ReCTS~\citep{liu2019rects}, VinText~\citep{nguyen2021vintext}, MLT-2019~\citep{nayef2019mlt}, ThaiOCRBench~\citep{thaiocrbench2025} & 11{,}380 \\
Printed document & OmniDocBench~\citep{omnidocbench2025}, VieBookRead~\citep{do2025viebookread}, RETAS~\citep{yalniz2011retas}, Mozhi~\citep{mathew2024mozhi} & 113{,}118 \\
\midrule
Total & 25 datasets, 16 languages & 229{,}395 \\
\bottomrule
\end{tabular}
\caption{\textbf{The evaluation corpus contains 229,395 transcription lines after filtering.}}
\label{tab:corpus-summary}
\end{table*}

\begin{table}[!htb]
\centering\small
\setlength{\tabcolsep}{4pt}
\begin{tabular}{@{}lrl@{}}
\toprule
System & Params & Reference \\
\midrule
\multicolumn{3}{@{}l}{\textit{Classical OCR}}\\
\quad EasyOCR           & --      & \citep{easyocr} \\
\quad PP-OCRv5          & --      & \citep{ppocrv5} \\
\multicolumn{3}{@{}l}{\textit{Document-specialized VLM}}\\
\quad Qianfan-OCR       & 4B       & \citep{qianfanocr} \\
\multicolumn{3}{@{}l}{\textit{General VLMs}}\\
\quad Gemma 4           & 26B-A4B  & \citep{gemma4} \\
\quad InternVL3.5       & 38B      & \citep{internvl35} \\
\quad Qwen3.5           & 9B       & \citep{qwen35blog} \\
\quad Qwen3.6           & 27B      & \citep{qwen36_27b} \\
\bottomrule
\end{tabular}
\caption{\textbf{The evaluation includes classical pipelines, a document-specialized VLM, and general VLMs.} 26B-A4B gives the total and active parameters of a mixture-of-experts model.}
\label{tab:systems}
\end{table}

\subsection{Human studies}
\label{app:annotation}

\paragraph{Readers.} Readers were native or fluent speakers naive to the research, recruited through Prolific and by invitation and paid \$12 per hour, the platform's recommended rate. They saw no system names, judge output, or study hypothesis, and no author rated items. \Cref{tab:reliability} gives their agreement.

\paragraph{Consent.} Readers recruited through Prolific saw a study description naming the purpose (evaluating how well OCR systems preserve meaning), the task, and its length. All readers could stop at any time and consented by starting. No ethics board reviewed the study.

\paragraph{Data.} Readers judged transcriptions of public documents, and we asked nothing about them. We stored only pseudonymous rater IDs, ratings, optional rationales, and response times. Since the description did not mention publishing ratings, we release per-item aggregates, the mean rating or majority choice and the number of readers, and no individual responses, IDs, or rationales.

\paragraph{Items and sampling.} A single-edit or pair-level item shows the reference and the transcription as two blocks of text under the question \emph{How severe is this OCR error?} The five levels of \cref{tab:rubric} appear as buttons labeled by the table's descriptions. Single-edit items mark the differing span, while pair-level items show plain text, so readers find the errors themselves. Preference items show the reference and two outputs in random order, and readers pick one. Before rating, readers study worked examples and must answer four of five severity checks correctly.

\paragraph{Sampling.} The single-edit pools take near-equal thirds from three bands of the judge's severity, $[0,0.3)$, $[0.3,0.7)$, and $[0.7,1]$. The pair-level pools sample the five quintiles of unnormalized \cer{}, weighted toward low error.

\begin{table}[!htb]
\centering\footnotesize
\setlength{\tabcolsep}{3.5pt}
\begin{tabular}{@{}llcccc@{}}
\toprule
Study & Lang & Items & Readers & $\alpha$ & ICC(2,$k$) \\
\midrule
\multirow{3}{*}{single-edit}
  & English & 87 & 5 & 0.71 & 0.89 \\
  & Vietnamese & 80 & 4 & 0.67 & 0.87 \\
  & Chinese & 86 & 5 & 0.53 & 0.81 \\
\midrule
\multirow{3}{*}{pair-level}
  & English & 178 & 4 & 0.79 & 0.92 \\
  & Vietnamese & 173 & 4 & 0.65 & 0.88 \\
  & Chinese & 185 & 3 & 0.68 & 0.84 \\
\midrule
\multirow{3}{*}{preference}
  & English & 186 & 4 & 0.59 & 0.85 \\
  & Vietnamese & 169 & 4 & 0.70 & 0.90 \\
  & Chinese & 169 & 4 & 0.51 & 0.81 \\
\bottomrule
\end{tabular}
\caption{\textbf{The annotation studies reach moderate to high reliability.} $\alpha$ is Krippendorff's alpha~\citep{krippendorff2004} with ordinal weighting, and ICC(2,$k$)~\citep{shrout1979icc} is the reliability of the readers' mean, both over readers who rated at least half of a pool's items.}
\label{tab:reliability}
\end{table}

\FloatBarrier
\subsection{Statistical tests}
\label{app:stats}

Holm correction~\citep{holm1979} runs within four kinds of family.
\begin{enumerate}
\setlength{\itemsep}{0pt}
\item In the preference test, each comparison on each set of pairs forms a family of its three languages, for \semcer{}'s margins and for those scored with the judge's earlier severities, the distilled scorer, and the second judge (\cref{fig:forest,tab:second-judge}).
\item On the pair-level pool, the contrasts with \semcer{} form one family per language (\cref{tab:family}). The class charge lies outside it, since we added it after that test, so its row carries an uncorrected interval.
\item The $24$ equal-count contrasts form one family (\cref{tab:equal-cost-app}).
\item Each comparison of the severity-sorted list with a baseline forms a family of its three languages, on the top-of-list share for the best judge-free order and the type table and on the top $10\%$ for \cer{}'s charge (\cref{tab:grade-full}).
\end{enumerate}
We compute each partial correlation on ranks after regressing both the score and the readers' mean rating on the sampling band and the named baseline. The single-edit partials given the band alone and given the type table, the filter label, or both carry no Holm correction, since each answers a pre-specified question. The partial given the mode, added after those, is reported the same way.

\section{Per-Edit Validation}
\label{app:grade}

\subsection{Top-of-list measure}
\label{app:review}

\Cref{sec:grade-order} scores an order of the pool's edits by how early it places the edits readers rate most severe. For the list sorted by a charge $c$, let $G_f(c)$ be the mean reader rating summed over the edits above the cut $f$. Let $O_f$ be the same sum under the readers' own order, the most that many edits can reach, and $R_f$ its expectation under a random order. The share averages the normalized gain over five cuts,
\begin{equation}
  \mathrm{share}(c)=\frac{1}{5}\sum_{f\in\{0.1,\,0.2,\,0.3,\,0.4,\,0.5\}}\frac{G_f(c)-R_f}{O_f-R_f},
  \label{eq:share}
\end{equation}
so a random order scores $0$ and the readers' own order $1$, and edits that $c$ ties count at their group's mean rating. The top $10\%$ measure of \cref{tab:grade-full} is $G_{0.1}(c)/O_{0.1}$, the share of the damage in the readers' own top $10\%$ that the list's top $10\%$ catches. We fixed the measure and its comparisons with the best judge-free order and the type table before computing them and added the comparison with \cer{}'s charge later.

\subsection{Baselines}
\label{app:baselines}

An edit type that the type table of \cref{sec:grade-table} lacks falls back to the mean severity of its edit class, then of the language. The \emph{judge-free orders} sort the pool by the filter label, the real-word label of \cref{app:behavior}, or \cer{}'s per-edit charge, or by all three, each breaking the ties of the one before. The best judge-free order is the one of the four with the highest share. The \emph{class charge} maps the judge's most likely level to the three classes of severity-aware WER~\citep{wang2024swer}, level $0$ to OK, levels $1$ and $2$ to minor, and levels $3$ and $4$ to critical. It charges them the weights $0.2$, $0.6$, and $1$. The \emph{whole-line judgment} has the judge rate the whole transcription once and scales that severity by \cer{}.

\subsection{Full results}

\Cref{tab:grade-full} collects the per-edit results. A partial correlation controls for the severity band, and for the mode, the judge's most likely level, where named.

\begin{table}[!htb]
\centering\footnotesize
\setlength{\tabcolsep}{3pt}
\begin{tabular}{@{}lccc@{}}
\toprule
 & English & Vietnamese & Chinese \\
\midrule
\multicolumn{4}{@{}l}{\emph{Partial correlation within band}} \\
severity & 0.30\rlap{$^{*}$} & 0.33\rlap{$^{*}$} & 0.52\rlap{$^{*}$} \\
mode & 0.44\rlap{$^{*}$} & 0.23\rlap{$^{*}$} & 0.29\rlap{$^{*}$} \\
severity, given the mode & 0.13 & 0.24\rlap{$^{*}$} & 0.45\rlap{$^{*}$} \\
\midrule
\multicolumn{4}{@{}l}{\emph{Correlation on the whole pool}} \\
severity & 0.81 & 0.66 & 0.82 \\
mode & 0.85 & 0.63 & 0.77 \\
filter label & 0.79 & 0.69 & 0.79 \\
type table & 0.64 & 0.29 & 0.73 \\
\midrule
\multicolumn{4}{@{}l}{\emph{Top-of-list share}} \\
severity & 0.73 & 0.58 & 0.81 \\
best judge-free order & 0.51 & 0.48 & 0.49\rlap{$^{\dagger}$} \\
type table & 0.64 & 0.28 & 0.72 \\
\cer{}'s per-edit charge & 0.09 & 0.03 & $-0.13$ \\
\midrule
\multicolumn{4}{@{}l}{\emph{Top $10\%$, share of the readers' own top $10\%$}} \\
severity & 0.82 & 0.81 & 0.92 \\
best judge-free order & 0.66 & 0.77 & 0.66 \\
\cer{}'s per-edit charge & 0.29\rlap{$^{\dagger}$} & 0.55\rlap{$^{\dagger}$} & 0.42\rlap{$^{\dagger}$} \\
\bottomrule
\end{tabular}
\caption{\textbf{Per-edit results on the single-edit pools.} The first two blocks are Spearman correlations with readers' ratings, the mode being the judge's most likely level, and the last two the top-of-list measures of \cref{app:review}. An asterisk marks a within-band partial whose interval excludes zero, and a dagger a baseline that severity leads significantly.}
\label{tab:grade-full}
\end{table}

The judge's most likely level alone leaves $70$ to $81\%$ of the pairs of edits within a band tied, and an edit's exact substitution (over substitutions seen at least five times) explains only $39$ to $58\%$ of the variance in severity, so most of the charge comes from the line. Readers' own within-band reliability, split-half with the Spearman--Brown correction, caps a partial at roughly $0.81$, $0.85$, and $0.71$.

\section{Whole Transcriptions and Preferences}
\label{app:decision}

\subsection{Whole transcriptions}
\label{app:family}

On the pair-level pool, normalization matters more than how edits are weighted (\cref{tab:family}). Normalized \cer{} leads significantly in English and by $0.03$ in Vietnamese, and in Chinese no convention changes agreement much. The class charge trails \semcer{} by $0.03$ in English and Vietnamese and leads it by $0.01$ in Chinese, and each of these differences has an uncorrected interval that excludes zero.

\begin{table}[!htb]
\centering\footnotesize
\setlength{\tabcolsep}{5.5pt}
\begin{tabular}{@{}lccc@{}}
\toprule
Score & English & Vietnamese & Chinese \\
\midrule
unnormalized \cer{}       & $0.777^{*}$          & $0.775^{*}$          & $0.791$ \\
whitespace stripped       & $0.857$              & $0.807$              & $0.796$ \\
character filter          & $0.894$              & $0.822$              & $0.756$ \\
\cfcer{}                  & $0.925^{*}$         & $0.902$              & $0.757$ \\
\midrule
case-folded, punct.\ kept & $0.838^{*}$          & $0.862$              & $0.794$ \\
\midrule
\semcer{}                 & $0.890$              & $0.873$              & $0.763$ \\
\midrule
class charge              & $0.858^{\dagger}$ & $0.841^{\dagger}$ & $0.778^{\dagger}$ \\
\bottomrule
\end{tabular}
\caption{\textbf{Normalization drives agreement with readers over whole transcriptions.} Each entry is Spearman's $\rho$ with readers on the pair-level pool. The first four conventions are cumulative and apply before alignment. An asterisk marks a difference from \semcer{} that is significant after Holm correction (\cref{app:stats}). The class charge was added after that test, so its dagger marks an uncorrected difference from \semcer{} whose interval excludes zero.}
\label{tab:family}
\end{table}

\subsection{Preference test and its subsets}
\label{app:decision-test}

The test of \cref{sec:beyond-decision} uses the near-miss pairs among those with a strict reader majority. Before it, we fixed the criterion that \semcer{} agree with readers significantly more often than normalized \cer{} in at least two languages after Holm correction. \semcer{} leads significantly in English and Chinese, so the test meets its criterion (\cref{tab:decision-detail}).

\paragraph{Near-miss pairs.} A pair is a near miss when its higher-\cer{} output meets three conditions, which we defined from an earlier reading of the same pool. It (i)~has an edit beyond whitespace, case, punctuation, or symbols, (ii)~shares more than $0.80$ of its graphemes with the reference, and (iii)~keeps the reference's whole-number tokens and its count of negations. The overlap in (ii) is the multiset intersection over the larger multiset, after folding case and diacritics and dropping whitespace and punctuation.

\paragraph{Normalization-proof pairs.} After the test, we split the near-miss pairs by whether normalized \cer{} picks the same output as unnormalized \cer{}, a split that uses no severity, and call the pairs where it does \emph{normalization-proof pairs}.

\begin{figure*}[!htb]
\centering
\includegraphics[max width=\linewidth]{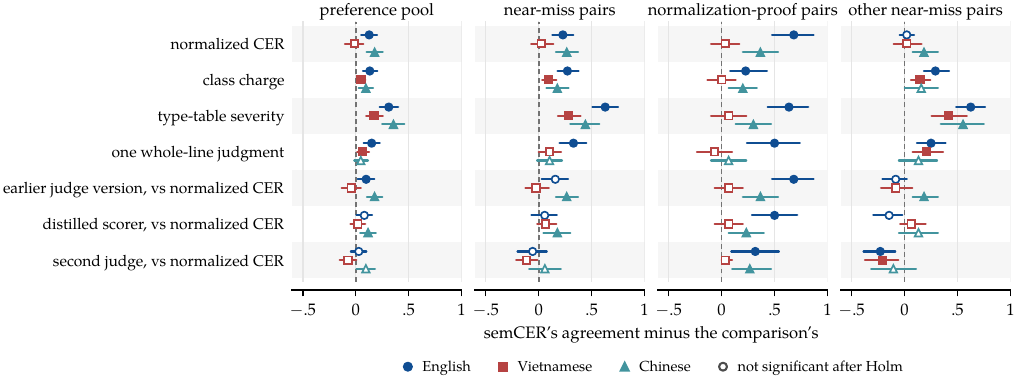}
\caption{\textbf{Every margin of the preference test.} Each marker is \semcer{}'s agreement with the readers' majority minus the comparison's, with its interval, a tie counting as a miss. Filled markers are significant. The last panel holds the near-miss pairs that are not normalization-proof, and the last three rows score the pairs with the judge's earlier severities, the distilled scorer, and the second judge.}
\label{fig:forest}
\end{figure*}

\paragraph{Baselines on this pool.}\label{app:holistic} The class charge rarely ties on the near-miss pairs (\cref{tab:decision-detail}), so ties do not explain its gap. The type table trails \semcer{} significantly on the whole pool in every language and agrees with readers less often on the near-miss pairs. The whole-line judgment trails \semcer{} on the preference pool, significantly in English and Vietnamese (\cref{fig:forest}).

\begin{table}[!htb]
\centering\footnotesize
\setlength{\tabcolsep}{4pt}
\begin{tabular}{@{}lccc@{}}
\toprule
 & English & Vietnamese & Chinese \\
\midrule
Pool, strict majority & 173 & 163 & 146 \\
Near-miss pairs & 70 & 78 & 68 \\
Normalization-proof pairs & 22 & 30 & 30 \\
\midrule
\multicolumn{4}{@{}l}{\emph{Agreement with readers on near-miss pairs}} \\
\semcer{} & 0.87\rlap{$^{*}$} & 0.76 & 0.85\rlap{$^{*}$} \\
\cfcer{} & 0.64 & 0.73 & 0.59 \\
class charge & 0.60 & 0.67 & 0.68 \\
\midrule
\multicolumn{4}{@{}l}{\emph{Near-miss pairs tied (\%)}} \\
class charge & 3 & 0 & 0 \\
\midrule
One reader with the rest & 0.92 & 0.91 & 0.94 \\
\bottomrule
\end{tabular}
\caption{\textbf{Counts behind the preference test.} An asterisk marks a significant margin over normalized \cer{}, and the last row is a single reader's agreement with the strict majority of the other readers.}
\label{tab:decision-detail}
\end{table}

\subsection{Robustness to the judge}
\label{app:robustness}

\paragraph{Judge version.}\label{app:judge-version} The provider replaced the judge's checkpoint during the study, and we drew the preference pool on the severities from before the change. With the severities we report, \semcer{} still ranks the two outputs opposite to \cer{} on $61$ of $70$, $47$ of $78$, and $48$ of $68$ near-miss pairs. The earlier severities charged every undecided edit one instead of asking the judge again. Under them, the English margin over normalized \cer{} on the near-miss pairs falls from $23$ to $16$ points and loses significance after Holm correction, the Vietnamese margin moves to $-3$ points, and the test misses its criterion. The difference lies in the near-miss pairs with an edit the judge first left undecided. The English and Chinese margins stay significant on the whole pool and on the normalization-proof pairs under both (\cref{fig:forest}).

\paragraph{Second judge.}\label{app:second-judge} Qwen3.7-Flash~\citep{qwen37flash} scores every rated transcription under the prompt of \cref{app:prompt}. Its severities correlate with the first judge's at $0.74$ or more, and it reproduces the per-edit and whole-transcription results (\cref{tab:second-judge}). The preference pool was drawn where the first judge overrides \cer{}, and the second judge overrides \cer{} on $42$ to $53\%$ of its pairs. Readers side with it on nine in ten or more of those overrides, and on the normalization-proof pairs its margin over normalized \cer{} is significant in English and Chinese, as the first judge's is (\cref{fig:forest}). On all near-miss pairs the margin is not significant, since the second judge sides with \cer{} on about half of them.

\begin{table}[!htb]
\centering\scriptsize
\setlength{\tabcolsep}{3pt}
\begin{tabular}{@{}lccc@{}}
\toprule
 & English & Vietnamese & Chinese \\
\midrule
\multicolumn{4}{@{}l}{\emph{Single edits}} \\
correlation with readers & 0.80 & 0.65 & 0.83 \\
partial within band & 0.37\rlap{$^{*}$} & 0.35\rlap{$^{*}$} & 0.57\rlap{$^{*}$} \\
same, given the filter label & 0.35\rlap{$^{*}$} & 0.20 & 0.51\rlap{$^{*}$} \\
correlation with the first judge & 0.81 & 0.74 & 0.83 \\
\midrule
\multicolumn{4}{@{}l}{\emph{Whole transcriptions}} \\
Spearman's $\rho$ with readers & 0.899 & 0.872 & 0.775 \\
gain over unnormalized \cer{} & 0.12\rlap{$^{*}$} & 0.10\rlap{$^{*}$} & $-0.02$ \\
\midrule
\multicolumn{4}{@{}l}{\emph{Preference pool}} \\
overrides of \cer{} & 92 & 68 & 63 \\
readers' agreement on them & 0.90 & 1.00 & 0.92 \\
margin, normalization-proof pairs & 0.32\rlap{$^{*}$} & 0.03 & 0.27\rlap{$^{*}$} \\
\bottomrule
\end{tabular}
\caption{\textbf{The second judge reproduces the per-edit, whole-transcription, and normalization-proof results.} Each entry is Qwen3.7-Flash's value on the rated pools. An override is a pair with a strict reader majority where it picks the output \cer{} rejects. The margin is its agreement with the readers' majority minus normalized \cer{}'s, a tie counting as a miss. An asterisk marks an interval that excludes zero, Holm-corrected in the margin row.}
\label{tab:second-judge}
\end{table}

\section{Corpus Analyses}

\subsection{Equal-count comparison}
\label{app:equal-cost}

\begin{table*}[t]
\centering\scriptsize
\setlength{\tabcolsep}{4pt}
\begin{tabular}{@{}lrrrrrr@{}}
\toprule
VLM & Chinese & Vietnamese & German & French & English & Spanish \\
\midrule
Gemma 4
  & $+0.093^{*}$ (456) & $+0.045$ (108) & $-0.010$ (479)
  & $-0.029^{*}$ (492) & $-0.011^{*}$ (1{,}431) & $-0.031^{*}$ (626) \\
InternVL3.5
  & $+0.028$ (343) & $+0.162^{*}$ (104) & $+0.002$ (260)
  & $+0.003$ (321) & $-0.010$ (1{,}242) & $+0.004$ (533) \\
Qwen3.5
  & $+0.051$ (308) & $+0.038$ (84) & $-0.006$ (473)
  & $-0.021^{*}$ (500) & $-0.011^{*}$ (1{,}395) & $-0.030$ (624) \\
Qwen3.6
  & $+0.012$ (297) & $+0.046$ (77) & $-0.016$ (506)
  & $-0.037^{*}$ (546) & $-0.010$ (1{,}309) & $-0.026$ (629) \\
\bottomrule
\end{tabular}
\caption{\textbf{Equal-count contrasts, each VLM against EasyOCR.} Each entry is the mean severity per edit, VLM minus EasyOCR, on
lines where both make the same number of edits, one to five, with paired lines in parentheses. An asterisk marks a contrast that survives Holm correction over the $24$.}
\label{tab:equal-cost-app}
\end{table*}

The pre-specified criterion for \cref{tab:equal-cost-app}, a gap above zero that survives Holm correction for every VLM in one language, holds in no language. In Chinese and Vietnamese every gap is positive, but only one per language survives correction, and the Vietnamese pairings rest on $77$ to $108$ lines each. In English every gap is negative, and in German, French, and Spanish the direction varies with the system. We fixed EasyOCR as the comparator and the six languages before the first run, since elsewhere the classical pipelines make more than five edits on nearly every line.

\subsection{Grapheme load}

\Cref{tab:grapheme-load} gives the values behind the language comparison of \cref{sec:corpus}.

\begin{table}[!htb]
\centering\footnotesize
\setlength{\tabcolsep}{5pt}
\begin{tabular}{@{}lccc@{}}
\toprule
Language & mark-only & content sub. & per edit \\
\midrule
Chinese$^{\dagger}$ & -- & 0.80 & 0.50 \\
Telugu & 0.75 & 0.81 & 0.74 \\
Kannada & 0.70 & 0.80 & 0.70 \\
Tamil & 0.66 & 0.77 & 0.68 \\
Thai & 0.64 & 0.81 & 0.66 \\
Marathi & 0.56 & 0.68 & 0.62 \\
Hindi & 0.52 & 0.73 & 0.44 \\
Vietnamese & 0.48 & 0.70 & 0.38 \\
German & 0.35 & 0.64 & 0.39 \\
Arabic & 0.33 & 0.81 & 0.55 \\
Russian & 0.31 & 0.74 & 0.60 \\
Portuguese & 0.28 & 0.76 & 0.62 \\
Urdu & 0.28 & 0.75 & 0.54 \\
French & 0.17 & 0.63 & 0.35 \\
English & 0.09 & 0.67 & 0.26 \\
Spanish & 0.06 & 0.66 & 0.42 \\
\bottomrule
\end{tabular}
\caption{\textbf{Severity per edit follows the grapheme load.} The judge's mean severity of a mark-only edit (the grapheme load), a content substitution, and all edits, over lines with one to five edits. Chinese$^{\dagger}$ has no marks, so its load is the content-substitution cost.}
\label{tab:grapheme-load}\label{tab:severity-per-edit}
\end{table}

\subsection{Real-word errors}
\label{app:behavior}

We assign each edit to its reference word, rebuild the word as the system wrote it, and call the result a real word when it occurs at least twice in the language's references. In Chinese, we use character pairs found in the references. Chance re-corrupts each erroneous word $20$ times with the same numbers of grapheme substitutions, insertions, and deletions, drawn from the language's letter frequencies.

The counts cover the ten languages of \cref{tab:chance}, those with a word list and at least $30$ letter errors in some domain. \Cref{fig:system} plots the same languages except Hindi, whose word list recognizes few words, and adds Chinese, for which the letter-frequency baseline is undefined. VLM letter errors form real words more often than chance predicts in every language and domain except Hindi scene text, while classical errors stay within $0.10$ of chance (\cref{tab:chance}).

\float{fig_system}

\begin{table}[!htb]
\centering\scriptsize
\setlength{\tabcolsep}{4pt}
\renewcommand{\arraystretch}{0.95}
\begin{tabular}{@{}llcc@{}}
\toprule
Language & Domain & general VLM & classical \\
\midrule
English & printed & $0.14$ / $0.05$\rlap{$^{*}$} & $0.18$ / $0.22$\rlap{$^{*}$} \\
English & handwriting & $0.57$ / $0.14$\rlap{$^{*}$} & -- \\
English & scene text & $0.22$ / $0.12$\rlap{$^{*}$} & $0.24$ / $0.20$\rlap{$^{*}$} \\
German & printed & $0.30$ / $0.03$\rlap{$^{*}$} & $0.04$ / $0.05$ \\
German & handwriting & $0.45$ / $0.04$\rlap{$^{*}$} & $0.05$ / $0.05$ \\
French & printed & $0.30$ / $0.04$\rlap{$^{*}$} & $0.07$ / $0.09$ \\
French & handwriting & $0.69$ / $0.11$\rlap{$^{*}$} & -- \\
Spanish & printed & $0.29$ / $0.02$\rlap{$^{*}$} & $0.11$ / $0.10$ \\
Portuguese & handwriting & $0.69$ / $0.13$\rlap{$^{*}$} & -- \\
Russian & handwriting & $0.40$ / $0.06$\rlap{$^{*}$} & -- \\
Vietnamese & printed & $0.87$ / $0.18$\rlap{$^{*}$} & $0.25$ / $0.20$\rlap{$^{*}$} \\
Vietnamese & handwriting & $0.93$ / $0.17$\rlap{$^{*}$} & -- \\
Vietnamese & scene text & -- & $0.40$ / $0.30$ \\
Arabic & handwriting & $0.22$ / $0.02$\rlap{$^{*}$} & -- \\
Urdu & printed & $0.55$ / $0.12$\rlap{$^{*}$} & $0.46$ / $0.47$ \\
Urdu & handwriting & $0.60$ / $0.12$\rlap{$^{*}$} & -- \\
Hindi & scene text & $0.08$ / $0.05$ & -- \\
\bottomrule
\end{tabular}
\caption{\textbf{General VLM letter errors form real words far above chance, while classical errors stay near it.} Each entry is the share of letter errors that form another real word, observed / chance. An asterisk marks a difference from chance whose interval excludes zero, and a dash fewer than $30$ letter errors.}
\label{tab:chance}
\end{table}

\FloatBarrier
\section{Distilled Scorer}
\label{app:student}

We split the lines outside the rated pools by reference line into training, validation, and test sets, train the student as in \cref{tab:student-setup}, and fix the number of training steps before testing. \Cref{tab:student-results} compares it with the judge. Its correlation with readers on the single-edit pools comes within $0.03$ of the judge's. Since a user sorts a line's errors by the student's charge, we also check that it orders them as the judge does. A clear pair is two edits whose judge severities differ by at least half a rubric step, and the student orders nine in ten as the judge does. Its severity differs from the judge's by a mean absolute $0.04$ per edit on the test lines.

\begin{table}[!htb]
\centering\footnotesize
\setlength{\tabcolsep}{4pt}
\begin{tabular}{@{}lcccc@{}}
\toprule
 & English & Vietnamese & Chinese & All $16$ \\
\midrule
\multicolumn{5}{@{}l}{\emph{Correlation with readers, single edits}} \\
student & 0.79 & 0.63 & 0.80 & \\
judge & 0.81 & 0.66 & 0.82 & \\
\midrule
\multicolumn{5}{@{}l}{\emph{Clear pairs ordered as the judge does (\%)}} \\
held-out test lines & 90 & 91 & 93 & 90 \\
rated-pool lines & 90 & 92 & 90 & 91 \\
\bottomrule
\end{tabular}
\caption{\textbf{The distilled scorer tracks readers and the judge.} The test lines are $1{,}623$ sampled from the test split, up to $120$ per language, each transcribed by one randomly chosen system. The pooled share is $90.3\%$ [$89.5$, $91.1$], and no language falls below $87\%$.}
\label{tab:student-results}
\end{table}

\begin{table}[!htb]
\centering\footnotesize
\setlength{\tabcolsep}{4pt}
\begin{tabular}{@{}ll@{}}
\toprule
Data & $192{,}897$ training, $11{,}344$ validation, \\
 & and $22{,}683$ test lines \\
Backbone & Qwen3.5-0.8B-Base~\citep{qwen35blog}, \\
 & all weights trained \\
Input & marked pair, up to $2{,}048$ tokens \\
Readout & last token and the pooled \\
 & reference and output spans \\
Head & MLP, $256$ hidden units, GELU, \\
 & dropout $0.1$, five rubric levels \\
Target & the judge's distribution over levels \\
Loss & KL divergence + ranked probability \\
 & score, balanced by gradient norm \\
Sampling & languages balanced, low severities \\
 & drawn above their corpus share \\
Optimizer & AdamW, $\beta=(0.9, 0.95)$, weight \\
 & decay $0.1$ on weight matrices \\
Learning rate & $10^{-5}$ backbone, $10^{-4}$ head \\
Schedule & warm-up $2$k, constant to $80$k, \\
 & linear decay to $0$ at $100$k steps \\
Batch, clipping & $32$ edits; gradient norm $1.0$ \\
\bottomrule
\end{tabular}
\caption{\textbf{Training setup of the distilled scorer.}}
\label{tab:student-setup}
\end{table}

\end{document}